\documentclass[letterpaper]{article} 
\usepackage{aaai24}  
\usepackage{times}  
\usepackage{helvet}  
\usepackage{courier}  
\usepackage[hyphens]{url}  
\usepackage{graphicx} 
\usepackage{natbib}  
\usepackage{caption} 
\usepackage{algorithm}
\usepackage{algorithmic}
\usepackage{newfloat}
\usepackage{listings}
\DeclareCaptionStyle{ruled}{labelfont=normalfont,labelsep=colon,strut=off} 
\floatstyle{ruled}
\newfloat{listing}{tb}{lst}{}
\floatname{listing}{Listing}
\usepackage{multirow} 
\usepackage{multicol}
\usepackage{color}
\usepackage{bm}
\usepackage{amsfonts}

\title{LF-ViT: Reducing Spatial Redundancy in Vision Transformer for Efficient \\ Image Recognition}
\author {
    Youbing Hu\textsuperscript{\rm 1}, 
    Yun Cheng\textsuperscript{\rm 2}\thanks{Corresponding Author},
    Anqi Lu\textsuperscript{\rm 1},
    Zhiqiang Cao\textsuperscript{\rm 1},
    Dawei Wei\textsuperscript{\rm 3},
    Jie Liu\textsuperscript{\rm 1},
    Zhijun Li\textsuperscript{\rm 1}\footnotemark[1]
}
\affiliations {
    \textsuperscript{\rm 1}AIoT Lab, Faculty of Computing, Harbin Institute of Technology\\
    \textsuperscript{\rm 2}Swiss Data Science Center, Zurich, Switzerland \\
    \textsuperscript{\rm 3}School of Cyber Engineering, Xidian University\\
    youbing@stu.hit.edu.cn, yun.cheng@sdsc.ethz.ch, luanqi@stu.hit.edu.cn\\ zhiqiang\_cao@stu.hit.edu.cn, weidawei58@gmail.com, \{jieliu, lizhijun\_os\}@hit.edu.cn
}

\usepackage{bibentry}

\begin{document}

\maketitle

\begin{abstract}

The Vision Transformer (ViT) excels in accuracy when handling high-resolution images, yet it confronts the challenge of significant spatial redundancy, leading to increased computational and memory requirements. To address this, we present the Localization and Focus Vision Transformer (LF-ViT). This model operates by strategically curtailing computational demands without impinging on performance. In the Localization phase, a reduced-resolution image is processed; if a definitive prediction remains elusive, our pioneering Neighborhood Global Class Attention (NGCA) mechanism is triggered, effectively identifying and spotlighting class-discriminative regions based on initial findings. Subsequently, in the Focus phase, this designated region is used from the original image to enhance recognition. Uniquely, LF-ViT employs consistent parameters across both phases, ensuring seamless end-to-end optimization. Our empirical tests affirm LF-ViT's prowess: it remarkably decreases Deit-S's FLOPs by 63\% and concurrently amplifies throughput twofold. Code
of this project is at https://github.com/edgeai1/LF-ViT.git.
\end{abstract}

\section{Introduction}

Transformer \cite{vaswani2017attention} is currently the most popular architecture in natural language processing (NLP) tasks, attracting the attention of an increasing number of researchers. Within the computer vision community, the success of the vision transformer (ViT) \cite{ViT} has been remarkable and its influence continues to expand. The transformer architecture is built upon the self-attention mechanism, enabling efficient capture of long-range dependencies between different regions in input images. This capability has led to the widespread application of transformers in tasks such as image classification \cite{chen2021crossvit, han2021transformer, li2021localvit, swin_transformer, Deit, wu2021cvt}, object detection \cite{carion2020end, dai2021up, sun2021rethinking}, and semantic segmentation \cite{li2022panoptic, wang2023dformer}.

The predominant ViT architecture relies predominantly on segmenting a 2D image into sequentially arranged patches and transforming these patches into one-dimensional tokens using linear mappings \cite{ViT}. Following this, the self-attention mechanism facilitates interactions between these tokens, thereby handling essential computer vision operations. However, the use of ViT for image analysis results in substantial computational overhead, which increases quadratically with the number of tokens \cite{swin_transformer}.
Although ViT exhibits outstanding efficacy during training and inference with high-resolution images \cite{Deit}, its computational demands surge considerably with rising input image resolutions. This notable increase hampers the viability of deploying ViT models on resource-limited edge devices and Internet of Things (IoT) systems \cite{ignatov2019ai}.

To optimize the computational efficiency of ViT, researchers have proposed several optimization strategies and extension methods \cite{xu2022evo, meng2022adavit, yin2022vit, cf_vit, wang2021not, peng2021conformer, chen2021visformer}. 
For example,
DVT \cite{wang2021not} cascades multiple ViTs with increasing tokens and then leverages an early exit policy to decide the token number of each image.  
CF-ViT \cite{cf_vit} introduces a two-stage network inference approach. In the coarse stage, the input image is divided into shorter patch sequences to enable computationally efficient classification. When recognition is inadequate, meaningful patches are pinpointed, and subsequently subject to finer partition during the fine stage.
However, most of these methods focus on excavating redundant tokens in ViT, without fully utilizing the inherent spatial redundancy of high-resolution images to reduce computational costs.

In this paper, we seek to reduce the computational cost introduced by high-resolution input images in ViT. Our motivation stems from the fact that not all regions in an image are task-relevant, resulting in significant spatial redundancy during image recognition. We train Deit-S (Touvron et al. 2021a) with images of different resolutions and report top-1 accuracy and FLOPs in Table~\ref{tab:tab1}, in which, with a 4.2$\times$ improvement in computational cost, the image resolution using 224$\times$224 is only obtained with an accuracy advantage of 6.5\%. 
This indicates that a large amount of spatial redundancy exists in the image. 
In fact, GFNet \cite{wang2020glance} has demonstrated that only a small portion of an image, such as the head of a dog or the wings of a bird, which possesses class-discriminative characteristics, is sufficient for accurate image recognition. These regions are typically smaller than the entire image, requiring fewer computational resources. Therefore, if we can dynamically identify the class-discriminative regions for each individual image and focus only on these smaller regions during the inference, we can significantly reduce spatial redundancy without sacrificing accuracy.
Therefore, to achieve the idea above, we need to address two key challenges: (1) how to efficiently identify class-discriminative regions with minimal area. (2) how to adaptively allocate computational resources to each individual image considering the varying number of class-discriminative regions may differ across different inputs.

\begin{table}[t]

\centering
\begin{tabular}{l|ll}
\hline
    Resolutions & 224$\times$224 & 112$\times$112 \\
    \hline
    Accuracy & 79.8\% & 73.3\%  \\
    FLOPs & 4.60G & 1.10G \\
\hline
\end{tabular}
\caption{Accuracy and FLOPs of Deit-S \cite{Deit} on ImageNet \cite{deng2009imagenet} with different image resolutions as input.}
\label{tab:tab1}
\end{table}

\begin{figure}[t]
\centering
\includegraphics[width=0.90\columnwidth]{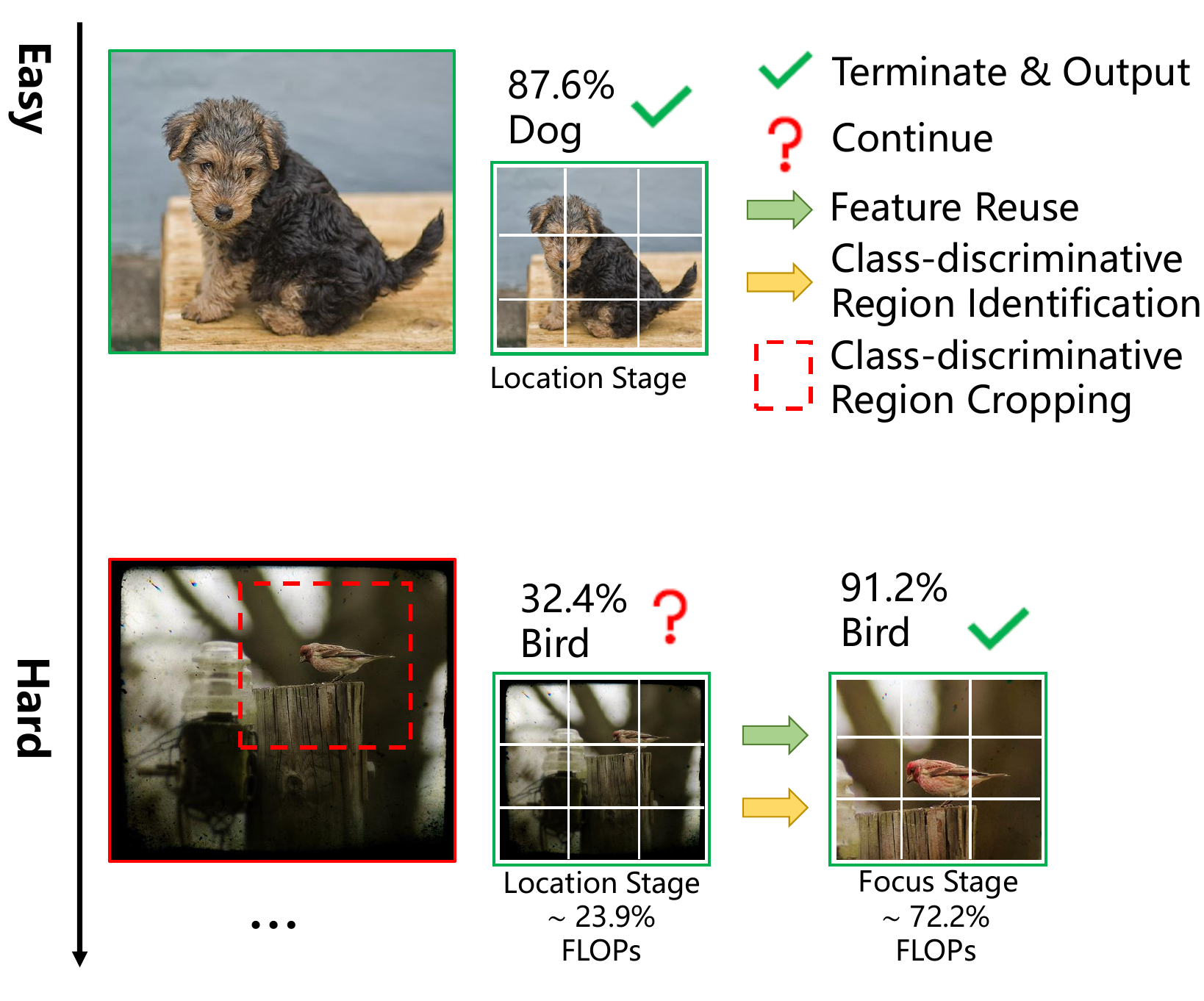}
\caption{Examples of LF-ViT. FLOPs refer to the proportion of the computation required by LF-ViT (e.g., down-sampled to 112$\times$112) versus processing the entire 224$\times$224 input image. 
}
\label{fig:fig1}
\end{figure}

In this paper, we introduce LF-ViT, a two-stage image recognition framework designed to address the aforementioned challenges. Our goal with LF-ViT is to produce accurate predictions while adaptively optimizing the computational cost of inputs. As depicted in Fig.~\ref{fig:fig1}, the inference process of LF-ViT consists of two stages: localization and focus.
During the localization stage, LF-ViT initiates inference on a down-sampled version of each image. If the resulting predictions are sufficiently confident, the inference concludes promptly, and the outcomes are presented. By operating on these down-sampled, low-resolution images, LF-ViT minimizes computational expenses, realizing efficiency gains. If the predictions aren't conclusive, the Neighborhood Global Class Attention (NGCA) mechanism leverages the outputs from the localization stage to pinpoint the class-discriminative regions in the full-resolution image.
Following this, we identify the class-discriminative regions from the tokens generated by the original image embeddings and select the top-K tokens with the most pronounced Global Class Attention (GCA) to hone in on image recognition during the focus stage. 

In addition, we introduce two mechanisms centered on computational reuse: the non-class-discriminative region feature reuse mechanism and the class-discriminative region feature fusion mechanism. The first mechanism repurposes features from non-class-discriminative regions and from non-top-K areas within class-discriminative regions identified during the localization stage. This repurposing provides essential background details, enhancing the accuracy of image recognition in the focus stage. Conversely, the second mechanism amalgamates features of class-discriminative regions from the localization stage with those from the focus stage, thereby amplifying LF-ViT's overall performance.

Both the localization and focus stages of LF-ViT utilize the same network parameters and are optimized jointly in an end-to-end approach. We put our proposed LF-ViT, which is built upon DeiT \cite{Deit}, to the test on ImageNet. Comprehensive experimental results demonstrate that LF-ViT significantly enhances inference efficiency. For instance, while maintaining an accuracy of 79.8\%, LF-ViT cuts down DeiT-S's FLOPs by 63\% and doubles the practical throughput to 2.03 times on an A100 GPU.

\section{Related Work}

\subsection{Vision Transformer}

Inspired by the great success of transformer \cite{vaswani2017attention} on NLP tasks, many recent studies have explored the introduction of transformer architecture to multiple computer vision tasks \cite{carion2020end, cf_vit, chen2021crossvit, chen2021visformer, li2021localvit, dai2021up, Deit, swin_transformer, wang2023dformer, wang2021not, meng2022adavit}. Following ViT \cite{ViT}, a variety of ViT variants have been proposed to improve the recognition performance as well as training and inference efficiency. DeiT \cite{Deit} incorporates distillation strategies to improve the training efficiency of ViTs, outperforming standard CNNs without pretraining on large-scale datasets like JFT \cite{sun2017revisiting}.  LV-ViT \cite{lv_vit} leverages all tokens to compute the training loss, and the location-specific supervision label of each patch token is generated by a machine annotator.  
GFNet \cite{rao2021global} replaces self-attention in ViT with three key operations that learn long-range spatial dependencies with log-linear complexity in the frequency domain. CrossViT \cite{chen2021crossvit} achieves new SOTA performance using a two-branch transformer to combine different patch sizes to recognize objects across multiple scales.
DeiT III \cite{touvron2022deit} boosts the supervised training performance of ViT models on ImageNet to a new benchmark by improving the training strategy.  
Wave-ViT \cite{yao2022wave} seeks a better trade-off between efficiency and accuracy by formulating reversible downsampling via wavelet transform and self-attentive learning. 
Spectformer \cite{patro2023spectformer} first uses Fourier transform operations to implement a frequency domain layer used to extract image features at shallow locations in the network. In this paper, we focus on improving the performance of a generic ViT backbone, so our work is orthogonal to designing efficient ViT backbones.

\subsection{Adaptive Inference in Vision Transformer}
The human brain processes visual information using hierarchical and varied attention scales, enriching its environmental perception and object recognition \cite{gupta2021visual,zhang2017foveated}. This mirrors the adaptive inference rationale, which leverages the significant variances within network inputs as well as the redundancy in network architectures to improve efficiency through instance-specific inference strategies.  
In particular, previous techniques applied to CNNs have investigated various approaches, such as modifying input samples \cite{wu2020dynamic, zuxuan2021coarse}, skipping network layers \cite{wu2018blockdrop, wang2018skipnet} and channels \cite{lin2017runtime, bejnordi2019batch}, as well as employing early exiting with a multi-classifier structure \cite{bolukbasi2017adaptive, huang2017multi, li2019improved}.  In recent studies, researchers have explored the use of adaptive inference strategies to improve the inference efficiency of ViT models \cite{wang2021not, cf_vit, xu2022evo, tang2022quadtree}. 
Some studies \cite{yin2022vit, meng2022adavit, xu2022evo, rao2021dynamicvit} attempt to prune unimportant tokens dynamically and progressively during inference.  
DVT \cite{wang2021not} endows a proper token number for each input image by cascading three transformers. 
CF-ViT \cite{cf_vit} performs further fine-grained partitioning of informative regions scattered throughout the image to improve ViT model performance.
Compared to the aforementioned methods, our LF-ViT ingeniously harnesses the inherent spatial redundancy within images. By pinpointing and focusing on regions of class-discriminative within high-resolution images, we approach the issue from the perspective of spatial redundancy in images, effectively reducing the computational costs of the ViT model.

\section{Preliminaries}

Vision Transformer (ViT) \cite{ViT} splits images into sequences of patches as input, and then uses multiple stacked multi-head self-attention (MSA) and feed-forward network (FFN) building blocks to model the long-range dependencies between them. Formally, for each input image $\bm I^{C\times H \times W}$, ViT first splits into 2D patches with fixed size $\mathbf{X} = [\mathbf{x}_1,\mathbf{x}_2,...,\mathbf{x}_N]$, where $N$ is the number of patches, $C$, $H$, and $W$ denote the channel, height and width of the input image, respectively. These patches are then mapped to $D$-dimensional patch embeddings $\mathbf{Z} = [\mathbf{z}_1,\mathbf{z}_2,...,\mathbf{z}_N]$ with a linear layer, i.e., tokens. Subsequently, a learnable class token $\mathbf{z}_{cls}$ is appended to the tokens serving as a representation of the whole image. The positional embedding $\mathbf{E}_{pos}$ is also added to these tokens to enhance their positional information. Thus, the sequence of tokens input to the ViT model is:

\begin{equation}
    \mathbf{Z} = [\mathbf{z}_{cls};\mathbf{z}_1,\mathbf{z}_2,...,\mathbf{z}_N] + \mathbf{E}_{pos}
\end{equation}
where $\mathbf{z}\in \mathbb{R}^D$ and $\mathbf{E}_{pos}\in \mathbb{R}^{(N+1)\times D}$ respectively.

The backbone network of a ViT model consists of $L$ building blocks, each of which consists of a MSA and a FFN. In particular, the $l$-th encoder in a single-head, the token sequence $\mathbf{Z}_{l-1}$ is projected into a query matrix $\mathbf{Q}_l \in \mathbb{R}^{(N+1)\times D}$, a key matrix $\mathbf{K}_l \in \mathbb{R}^{(N+1)\times D}$, and a value matrix $\mathbf{V}_l \in \mathbb{R}^{(N+1)\times D}$. Then, the self-attention matrix $\mathbf{A}_l \in \mathbb{R}^{(N+1)\times(N+1)}$ is computed as:

\begin{equation}
    \mathbf{A}_l = {\rm Softmax}(\frac{\mathbf{Q}_l\mathbf{K}^T_l}{\sqrt{D}})\mathbf{V}_l = [\mathbf{a}_{cls,l};\mathbf{a}_{1,l},\mathbf{a}_{2,l},...,\mathbf{a}_{N,l}]\mathbf{V}_l
\end{equation}

The $\mathbf{a}_{cls,l} \in \mathbb{R}^{(N+1)}$ is known as class attention, reflecting the interactions between class tokens and other patch tokens. For more effective attention to different
representation subspaces, multi-head self-attention
concatenates the output from several single-head attentions
and projects it with another parameter matrix:

\begin{figure*}[t]
\centering
\includegraphics[width=1.75\columnwidth]{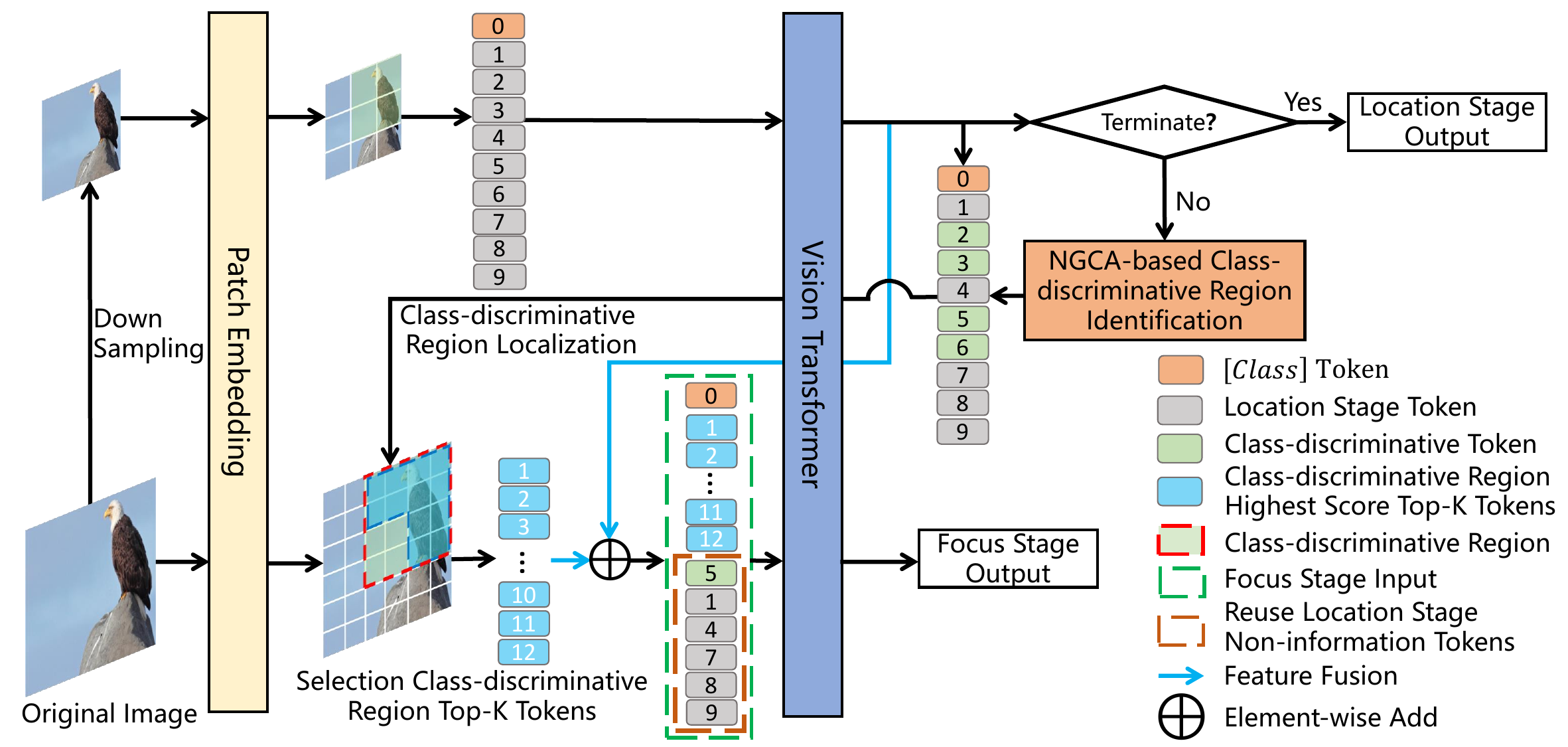}
\caption{
Overview of LF-ViT: (1) Input images are down-sampled and embedded using a consistent patch embedding for both the down-sampled and original image. (2) The down-sampled image undergoes ViT processing for localization. (3) If localization lacks a confident prediction, the Neighborhood Global Class Attention (NGCA) mechanism pinpoints class-discriminative regions in the original image. (4) The top-K tokens with peak global class attention (GCA) from these regions are used for focused recognition. Feature fusion and token reuse mechanisms optimize computation in the focus stage.
}
\label{fig:fig2}
\end{figure*}

\begin{equation}
    \mathbf{head}_{i,l} = \mathbf{A}(\mathbf{Z}_l\mathbf{W}^Q_{i,l}, \mathbf{Z}_l\mathbf{W}^K_{i,l}, \mathbf{Z}_l\mathbf{W}^V_{i,l})
\end{equation}
\begin{equation}
    {\rm MSA}(\mathbf{Z}_l) = {\rm Concat}(\mathbf{head}_{i,l},...,\mathbf{head}_{H,l})\mathbf{W}^O_l
\end{equation}
where $\mathbf{W}^Q_{i,l}$, $\mathbf{W}^K_{i,l}$, $\mathbf{W}^V_{i,l}$, $\mathbf{W}^O_l$ are the parameter matrices in the $i$-th attention head of the $l$-th build block, and $\mathbf{Z}_l$ denotes the input at the $l$-th block. The output from MSA is then fed into FFN to produce the output
of the build block $\mathbf{Z}_{l+1}$. Residual connections are also applied on both MSA and FFN as follows:
\begin{equation}
    \mathbf{Z}^{\prime}_l = {\rm MSA}(\mathbf{Z}_l) + \mathbf{Z}_l, \quad \mathbf{Z}_{l+1} = {\rm FFN}(\mathbf{Z}^{\prime}_l) + \mathbf{Z}^{\prime}_l
\end{equation}
The final prediction is produced by the classifier taking the class token $\mathbf{z}_{cls, L}$ from the last build block as inputs.

\section{Location and Focus Vision Transformer}
In this section, we describe our LF-ViT in detail. Our motivation comes from the fact that not all regions in an image are task-relevant. This inspires us to implement a localized and focused two-stage ViT, aiming to improve the computational efficiency of ViTs by performing focus computation on the minimal image regions to obtain a reliable prediction. As shown in Fig.~\ref{fig:fig2},  a down-sampled copy of the input image is used for image recognition in the localization stage.  If the image in the localization stage fails to obtain a convincing prediction, the class-discriminative region is selected from the original image to further focus recognition in the focus stage. Details are given below.

\subsection{Location Inference Stage}
LF-ViT first performs inference on a down-sampled copy of the input image $\bm I$ (down-sampled half to $H/2\times W/2$) to recognize ``easy'' images. It also localizes class-discriminative regions to achieve efficient image recognition when ``hard'' input images are encountered. In the localization stage, the input to LF-ViT is the vector $\mathbf{Z}$ (Eq.~1), then the output vector $\mathbf{Z}_L$ obtained after $L$ stacked MSA-FFN encoder transformations. The class token $\mathbf{z}_{cls, L}$ is input to the final classifier head $\mathcal{F}$ to obtain the final category prediction distribution $\mathbf{p}$:

\begin{equation}
    \mathbf{p} = \mathcal{F}(\mathbf{z}_{cls,L}) = [p_1, p_2, ... ,p_n]
\end{equation}
\begin{equation}
    j = \mathop{\rm arg\ max}\limits_{i} p_i
\end{equation}
where $n$ denotes the category number.
$p_j$ is the final prediction confidence score obtained in the localization stage, which we then compare with a pre-defined threshold $\eta$. If $p_j > \eta$, the inference process terminated immediately and $p_j$ is used as the output of the final prediction, which predicts the class $j$. Otherwise, the input image may be a ``hard'' image and class-discriminative regions need to be further localized to focus recognition. Note that the confidence threshold $\eta$ realizes a trade-off between performance and computation for our LF-ViT.

\textbf{Class-discriminative Regions Identification.} We revisit Eq. (1) where we add $\mathbf{z}_{cls}$ as a representation of the whole image, and then we transform it by the $l$-th MSA to get $\mathbf{a}_{cls,l}$, which represents the interaction between the class token and all image tokens. 
Therefore, we can use $\mathbf{a}_{cls,l}$ as a score to represent the importance of each image token in the $l$-th layer, similarly done in several ViTs optimization works \cite{cf_vit, liang2022evit, xu2022evo}. To more accurately and stably represent the importance of each token, instead of using the individual class attention that passes through the output of the last $L$-th MSA, we use the moving average class attention of each MSA in the entire ViT model instead of it. Formally, the global moving average class attention is as follows:

\begin{figure*}[t]
\centering
\includegraphics[width=1.80\columnwidth]{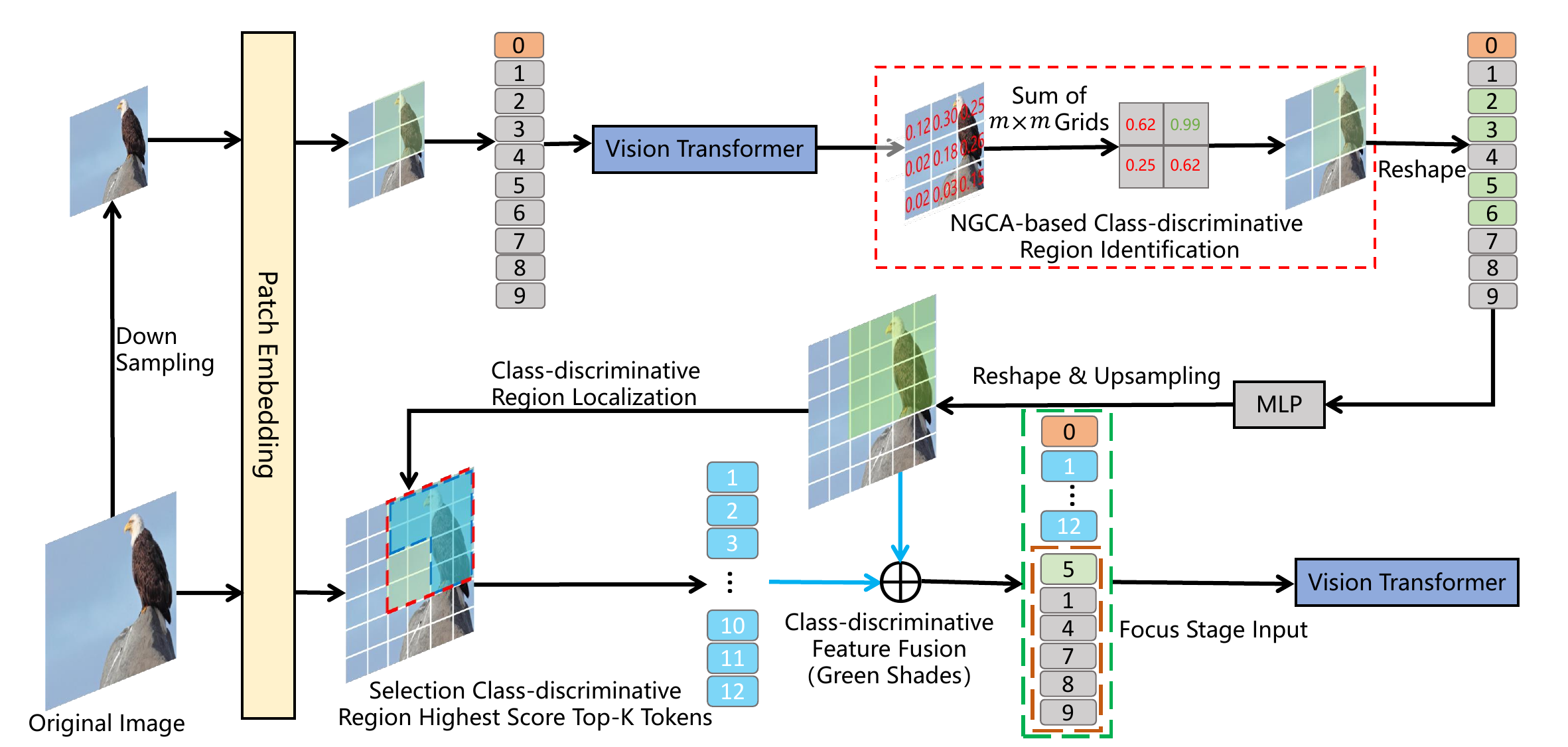}
\caption{ Illustration of our LF-ViT class-discriminative region identification, localization and feature fuse. The red numbers indicate the global class attention (GCA) of tokens. The green number indicates the region with the maximum neighborhood global class attention (NGCA), and we will select its corresponding region as the class-discriminative region. 
}
\label{fig:fig3}
\end{figure*}

\begin{equation}
    \mathbf{\overline{a}}_{cls,l} = \beta \cdot  \mathbf{\overline{a}}_{cls,l-1} + (1 - \beta) \cdot  \mathbf{a}_{cls,l}
\end{equation}
where $\beta$ = 0.99 and $l>2$. We select patches with high-score global
class attention (GCA) in the last encoder $\mathbf{\overline{a}}_{cls, L}$.
We intuitively argue that the tokens surrounding the tokens with the high-score GCA are also important for correctly recognizing images.
Motivated by this we propose neighborhood global class attentions (NGCA) to identify class-discriminative regions. 
As shown in the upper right corner of Fig.~\ref{fig:fig3}, we employ a grid of size $m \times m$ to compute the NGCA for each region in a sliding window manner on the GCA $\mathbf{\overline{a}}_{cls, L}$:

\begin{equation}
    \mathbf{a}_{r}  =  \mathcal{G}(\sum\limits_{i=1}^k \mathbf{\overline{a}}^{\, i}_{cls, L}) = [a_{r, 1},..., a_{r, M}]
\end{equation}
\begin{equation}
    g = \mathop{\rm arg\ max}\limits_{i} a_{r, i}
\end{equation}
where $m$ denotes the size of the region, $k$ denotes the tokens number in the region $m\times m$, $M$ denotes the total number of regions, and $\mathcal{G}$ denotes the computation of NGCA for M regions.
Finally, we select the $g$-th region with the highest GCA from all NGCA $\mathbf{a}_r$ serve as the class-discriminative region (Eq.~10).
As illustrated in Table~6, our NGCA mechanism adeptly pinpoints class-discriminative regions.

Our class-discriminative region identification method differs from earlier studies \cite{cf_vit, liang2022evit, xu2022evo} that used the GCA mechanism to indicate the importance of individual tokens, reducing computational complexity by minimizing token redundancy. In contrast, we employ the NGCA method to identify the class-discriminative regions from high-resolution images, thereby reducing model computational complexity from a spatial redundancy perspective. 
A comprehensive visualization can be found in Fig.~7.

\subsection{Focus Inference Stage}
When LF-ViT predicts the result $p_j<\eta$ in the localization stage, it indicates that a ``hard'' image is encountered that needs to perform the focus stage.

Upon determining the class-discriminative regions via the NGCA mechanism in Eq.10, it's crucial to locate these regions' token representations among the original image tokens. As depicted in Fig.\ref{fig:fig3}, for spatial alignment of the features from the last MSA-FFA layer with original image tokens, we first reshape the features. An MLP layer aids in flexible transformations, followed by upsampling and reshaping for dimension alignment. This upsampling doubles the tokens in the class-discriminative region, enabling direct location of these regions using the index from the aligned spatial dimension, as illustrated in Eq.~11:

\begin{equation}
    \mathbf{F}^{\prime} = {\rm Reshape}({\rm MLP}(\rm Upsample({\rm Reshape}(\mathbf{Z}_L))))
\end{equation}
where $\mathbf{F}^{\prime}\in HW/P^2 \times D$, $P$ denotes patch size. To further reduce the computational cost of LF-ViT, we introduce a threshold value $\alpha$ ($\alpha$ defaults to 0.88) to select the top-K most informative tokens from the class-discriminative regions. Specifically, we sort the class-discriminative regions based on the GCA obtained in the localization stage. Then, we select the highest GCA $\alpha \times m^2$ tokens from the localized class-discriminative regions in the original image as the input of the focus stage, while the remaining tokens are obtained by reusing the computation from the localization stage. Further details on token reuse will be discussed in the next section.

\textbf{Non-class-discriminative Regions Feature Reuse and Class-discriminative Regions Feature Fusion.} 
To provide necessary background information for target recognition during LF-ViT's focus stage, we reuse the non-class-discriminative regions identified in the localization stage, as well as the class-discriminative regions that are not included in $\alpha \times m^2$, to compensate for the lack of background information. These regions' tokens are represented by the gray dashed line below Fig.~\ref{fig:fig3}. Thus, we append them to the input token sequence during the focus stage.
As shown by the black arrows in Fig.~\ref{fig:fig3}, we perform element-wise fusion by adding the features from the class-discriminative regions identified in the localization stage and the class-discriminative regions localized in the original image (green shades). This fusion process enhances the semantic information of the input focus stage features, which improves the performance of LF-ViT. 
In Fig.~\ref{fig:fig5}, it's evident that our method of reusing features from non-class-discriminative regions and fusing features from class-discriminative regions greatly enhances the efficiency of LF-ViT.

\subsection{Training Objective}

During the training process of LF-ViT, we always set the confidence threshold $\eta$ = 1, which means that all images need to go through the inference of the focus stage. Following \cite{cf_vit}, our training objective for LF-ViT is two-fold. We aim for the output of the focus stage to be as consistent as possible with the ground truth labels, while also seeking the output of the localization stage to be similar to the output of the focus stage. Therefore, the training objective of LF-ViT can be summarized as follows:

\begin{equation}
    \mathcal{L}_{loss} = CE(\mathbf{p}_f; \mathbf{y}) + KL(\mathbf{p}_l; \mathbf{p}_f)
\end{equation}
where $CE(\cdot; \cdot)$ and $KL(\cdot; \cdot)$ respectively represent the cross entropy loss and Kullback-Leibler divergence. $\mathbf{p}_l$, $\mathbf{p}_f$, and $\mathbf{y}$ respectively represent the outputs of the localization stage,  the outputs  of the focus stage, and the ground truth labels.

\section{Experiments}

\subsection{Implementation Details}

We evaluate our LF-ViT on the ImageNet \cite{deng2009imagenet} image classification task, which is built on Deit-S \cite{Deit}. All our LF-ViTs use a patch size of 16$\times$16 to partition the images. To show the advantages of our
LF-ViT, we implement our method at two image resolutions on ImageNet, 224 and 288, denoted by LF-ViT and LF-ViT$^*$, respectively. For LF-ViT, the resolution of the input image is 224$\times$224, and we down-sample to 112$\times$112 in the localization stage, resulting in a total of 7$\times$7 tokens. For LF-ViT$^*$, the resolution of the input image is 288$\times$288, and we down-sample to 144$\times$144 in the localization stage, resulting in a total of 9$\times$9 tokens.

All the training strategies, such as data augmentation, regularization and optimizer, strictly follow the original settings of DeiT. We train LF-ViT for a total of 350 epochs. To improve performance and accelerate convergence, we recognize the entire image during the focus stage of the first 230 epochs, without class-discriminative region identification and localization. Only in the remaining epochs, the class-discriminative region identification is executed. Our LF-ViT always shares the same parameters in the localization and focus stages, including the parameters of patch embedding and ViT.

\begin{table*}[t]

\centering
\begin{tabular}{ccccc}
\hline
    \multirow{2}{*}{ Model}  &\multirow{2}{*}{$\eta$} & Acc. & FLOPs & Throughput  \\
     &  & (\%) & (G) & (img./s) \\
    \hline
    DeiT-S & - &79.8 &4.6 & 2601  \\
    \hline
    LF-ViT(4)  &  0.62 &79.8\color{black}{(+0.0)} & 1.8 \color{black}{($\downarrow$61\%)} & 4960\color{black}{($\uparrow$1.91$\times$)} \\
    LF-ViT(4)  &  0.77 &80.1\color{black}{(+0.3)} & 2.2 \color{black}{($\downarrow$52\%)} & 4415\color{black}{($\uparrow$1.70$\times$)} \\
    LF-ViT(5)  &  0.47 &79.8\color{black}{(+0.0)} & 1.7 \color{black}{($\downarrow$63\%)} & 5271\color{black}{($\uparrow$2.03$\times$)} \\
    LF-ViT(5)  &  0.61 &80.5\color{black}{(+0.7)} & 2.0 \color{black}{($\downarrow$57\%)}  & 4637\color{black}{($\uparrow$1.78$\times$)}\\
    LF-ViT(5)  &  0.76 &80.8\color{black}{(+1.0)} & 2.5 \color{black}{($\downarrow$46\%)} & 3686\color{black}{($\uparrow$1.42$\times$)}\\
    LF-ViT(6)  &  0.46 &79.8\color{black}{(+0.0)} & 1.8 \color{black}{($\downarrow$61\%)} & 5210\color{black}{($\uparrow$2.00$\times$)}\\
    LF-ViT(6)  &  0.65 &80.8\color{black}{(+1.0)} & 2.4 \color{black}{($\downarrow$48\%)} & 3764\color{black}{($\uparrow$1.45$\times$)}\\
    LF-ViT(6)  &  0.85 &81.0\color{black}{(+1.2)} & 3.0 \color{black}{($\downarrow$35\%)} & 2889\color{black}{($\uparrow$1.11$\times$)}\\
    \hline

\end{tabular}
\caption{Comparison between LF-ViT and its backbones. The first column shows the size of the class-discriminative region $m$.}
\label{tab:tab2}
\end{table*}

\subsection{Experimental Results}
\textbf{Model Performance.} To demonstrate the performance of our LF-ViT, we first compare LF-ViT with its backbone. We measure top-1 accuracy, model FLOPs, and model throughput as metrics. Following
existing studies \cite{cf_vit, liang2022evit}, the model throughput is measured as the number of processed images per second on a single A100 GPU. We feed the model 50,000 images in the validation set of ImageNet with
a batch size of 1024 and record the total inference time $T$. Then, the throughput is computed as $50,000 / T$.

Table \ref{tab:tab2} shows the comparison results with various thresholds $\eta$ and class-discriminative region sizes $m$. From the table, we can observe that when LF-ViT maintains the same accuracy as its backbone model, it significantly reduces Deit's FLOPs by 63\%, resulting in a maximum throughput improvement of 2.03$\times$. LF-ViT's outstanding performance is attributed to the design of our class-discriminative region identification and localization mechanism, which allows it to focus on the class-discriminative regions with the minimum area during the focus stage, thereby significantly improving efficiency. Furthermore, we have also discovered that increasing the value of $\eta$ significantly improves accuracy when $m$ remains the same. For instance, when $m$ = 6, $\eta$ increases from 0.46 to 0.85, LF-ViT enhances the accuracy of Deit-S by 1.2\%. These results convincingly demonstrate LF-ViT's ability to achieve a better trade-off between model accuracy performance and computational efficiency.

\begin{table}[t]

\centering
\begin{tabular}{ccc}
\hline
    Model &  Acc.(\%) & FLOPs(G) \\

    
    \hline
    Baseline\cite{Deit} & 79.8 & 4.6 \\
    DynamicViT\cite{rao2021dynamicvit} & 79.3 & 2.9 \\
  IA-RED$^{2}$  \cite{pan2021ia} & 79.1 & 3.2 \\
    PS-ViT\cite{psvit} &79.4 & 2.6 \\
    EViT \cite{liang2022evit} &  79.5&3.0 \\
    Evo-ViT\cite{xu2022evo} & 79.4& 3.0 \\
    A-ViT-S \cite{yin2022vit} & 78.6& 3.6 \\
    PVT-S\cite{wang2021pyramid} & 79.8 & 3.8 \\
    SaiT-S \cite{li2022sait} & 79.4 & 2.6 \\
    CF-ViT\cite{cf_vit} & 80.8& 4.0 \\
    CF-ViT$^*$\cite{cf_vit} & 81.9& 4.8 \\
    
    \hline
    \textbf{LF-ViT($m=5, \eta = 0.47$)} & \textbf{79.8} & \textbf{1.7} \\
    \textbf{LF-ViT($m=6, \eta = 0.65$)} & \textbf{80.8} & \textbf{2.4} \\
    \textbf{LF-ViT$^*$($m=8, \eta = 0.75$)} & \textbf{82.2} & \textbf{3.7} \\
\hline
\end{tabular}
\caption{Comparisons between existing token slimming-based ViT compression methods and our LF-ViT. $*$ denotes the coarse-grained stage of CF-ViT and the localization stage of LF-ViT with an input resolution of $144\times 144$.}
\label{tab:tab3}
\end{table}

\begin{figure}[t]
\centering
\includegraphics[width=1.0\columnwidth]{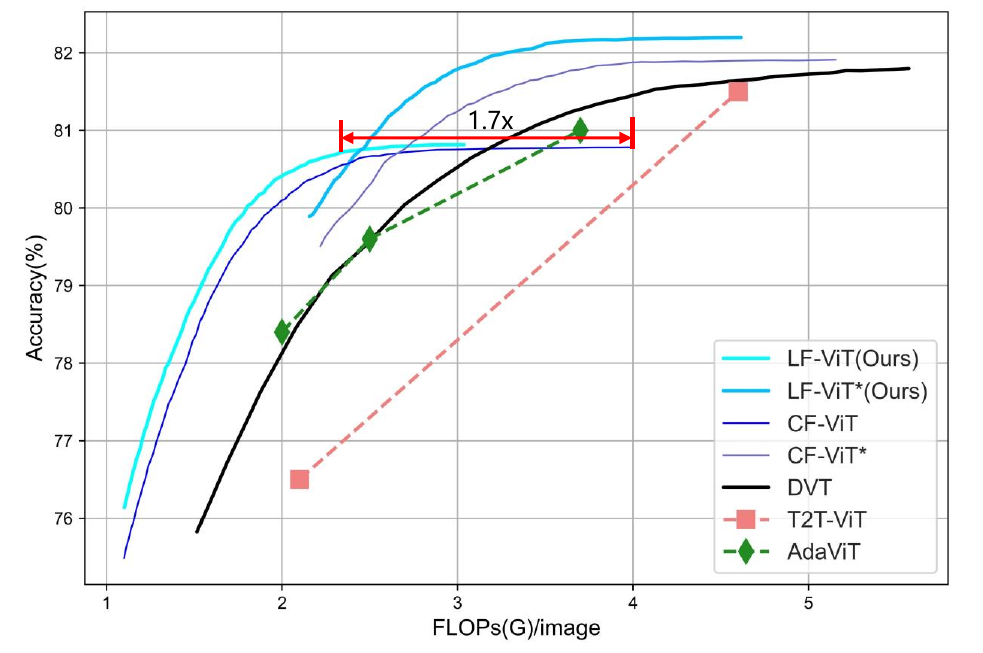}
\caption{Comparison between our LF-ViT and existing early-exiting methods. LF-ViT
obtains good efficiency/accuracy tradeoffs compared with other ViTs. DVT \cite{wang2021not}, CF-ViT \cite{cf_vit} and our LF-ViT are built upon DeiT.}
\label{fig:fig4}
\end{figure}

\textbf{Comparison with SOTA ViT Optimization Models.} To validate the efficiency of LF-ViT in reducing model complexity, we compare it with SOTA ViT optimization models, including token slimming compression and early-exiting compression.

(1) Token slimming compression reduces the complexity of the ViT model by progressively removing the number of input tokens, which is also the focus of this paper. Table~\ref{tab:tab3} shows the comparison between LF-ViT and these token slimming compression methods, including DynamicViT\cite{rao2021dynamicvit}, IA-RED$^2$ \cite{pan2021ia}, PS-ViT\cite{psvit}, EVIT \cite{liang2022evit}, Evo-ViT\cite{xu2022evo},  A-ViT-S \cite{yin2022vit}, PVT-S\cite{wang2021pyramid}, CF-ViT\cite{cf_vit} and SaiT-S \cite{li2022sait}. We report top-1 accuracy and FLOPs for performance evaluation. The results indicate that our LF-ViT outperforms these SOTA methods in terms of both accuracy improvement and FLOPs reduction. For instance, when achieving the same accuracy of 79.8\%, LF-ViT reduces FLOPs by 55\% compared to PVT-S. When the accuracy of 80.8\%, LF-ViT reduces FLOPs by 40\% compared to CF-ViT. Furthermore, when the coarse-grained stage of CF-ViT and the localization stage of LF-ViT with an input resolution of $144\times 144$, our LF-ViT$^*$ also achieves a 23\% reduction in FLOPs and a 0.3\% improvement in accuracy compared to CF-ViT$^*$.

\begin{figure}[t]
\centering
\includegraphics[width=0.95\columnwidth]{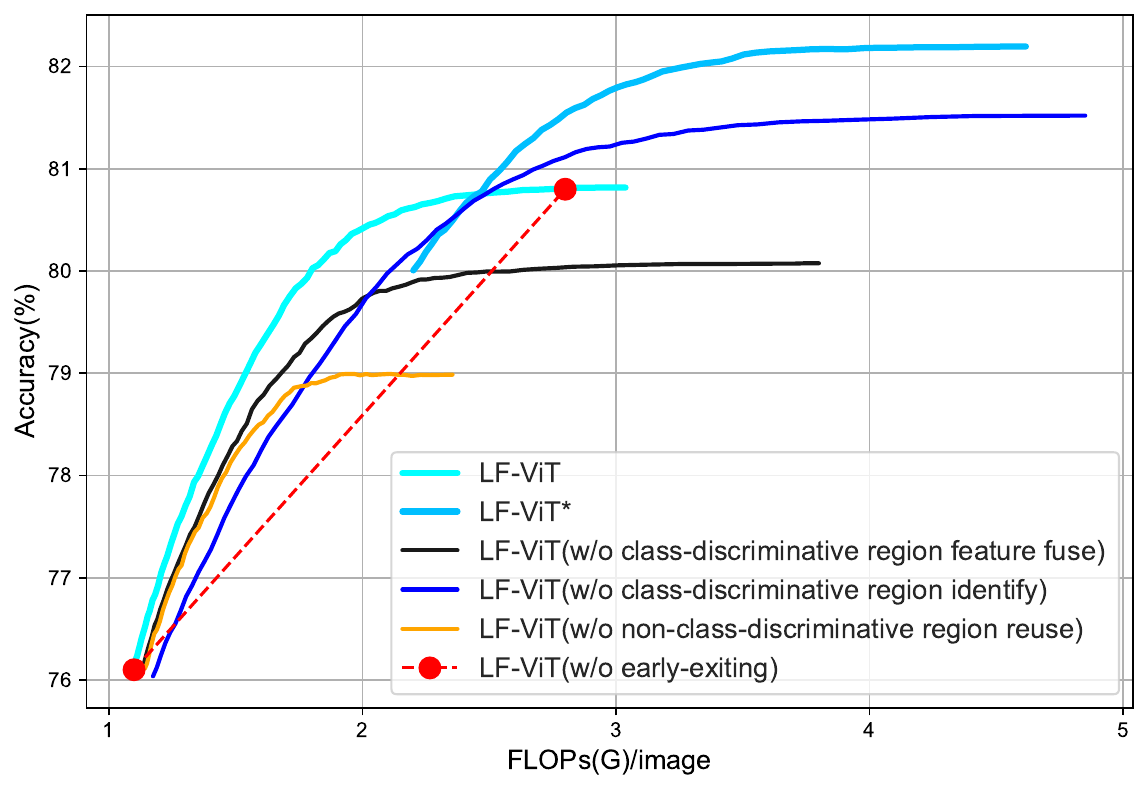}
\caption{Performance analysis of removing each of the four designs.}
\label{fig:fig5}
\end{figure}

(2) Early-exit compression halts the inference process if the intermediate representation of an input meets a specific exit criterion. This concept is also applied in the localization stage of our LF-ViT, where the computational graph stops if the prediction confidence $p_j$ exceeds the threshold $\eta$. Fig.~\ref{fig:fig4} shows the comparison between LF-ViT and these early-exit compression methods, including DVT \cite{wang2021not}, CF-ViT \cite{cf_vit}, T2T-ViT \cite{yuan2021tokens}, and AdapViT \cite{meng2022adavit}. From the figure we observed that our LF-ViT consistently outperforms all compared methods, achieving better trade-offs between accuracy and FLOPs. Compared to DVT cascading multiple ViT models with different numbers of input tokens, LF-ViT performs focus inference only in the class-discriminative region, greatly reducing the number of tokens. Compared to CF-ViT which performs further token splitting in discrete information regions distributed throughout the image, LF-ViT performs token splitting in the class-discriminative region with the minimum area, further reducing the number of tokens.  Thus, our LF-ViT improves efficiency by 1.7$\times$ without compromising accuracy. Even at higher image input resolutions, our LF-ViT$^*$ consistently outperforms CF-ViT$^*$ in terms of accuracy improvement and computational reduction.

In Fig.~7, we provide a detailed visual comparison between CF-ViT and LF-ViT.
In addition, our method is distinct from the token pruning-based techniques mentioned in \cite{liang2022evit, meng2022adavit, rao2021dynamicvit}. When combined, there's potential for enhanced efficiency. For instance, tokens from non-class-discriminative regions in LF-ViT can be processed with the EViT \cite{liang2022evit} approach to further diminish the token count.

\subsection{Ablation Study}
In this section, we analyze the efficiency of each design of LF-ViT individually, including class-discriminative regions identification, class-discriminative regions feature fusion, non-class-discriminative regions feature reuse, class-discriminative regions size, and early exit.

\begin{table}[t]

\centering
\begin{tabular}{ccccccc }
\hline
    $\beta$ & 0.0 & 0.5 & 0.9 & \textbf{0.99} & 0.999 \\
    \hline
    Acc.(\%) & 80.5 & 80.6 & 80.7 & 80.8 & 80.8 \\
    \hline

\end{tabular}
\caption{Accuracy with different values of $\beta$.}
\label{tab:tab4}
\end{table}

\begin{table}[t]

\centering
\begin{tabular}{ccccccc}
\hline
    $\alpha$ & 0.5 & 0.6 & 0.7 & 0.8 & \textbf{0.88} & 0.9 \\
    \hline
    Acc.(\%) & 80.1 & 80.2 & 80.6 & 80.6 &  80.8 & 80.7\\
    \hline

\end{tabular}
\caption{Accuracy with different values of $\alpha$.}
\label{tab:tab5}
\end{table}

\textbf{Influence of Class-discriminative Regions Size.} In Table~\ref{tab:tab2}, the numbers in the first column represent the size of the class-discriminative region, denoted as $m$. 
Smaller $m$ leads to fewer tokens and lower FLOPs in the focus stage but may result in decreased accuracy. In contrast, larger $m$ leads to more tokens and correspondingly higher accuracy in the focus stage. In all subsequent ablation studies, we conducted the experiments with $m$ = 5 and $\eta$ = 0.76 as default settings. 

\textbf{Necessity of Each Design.} Fig.~\ref{fig:fig5} plots the performance of our LF-ViT by individually removing each design. For LF-ViT*, the region size $m$ defaults to 8.
We can observe that removing each individual design of LF-ViT leads to a significant performance drop. These experiments demonstrate the critical importance of each design in achieving the superior performance of LF-ViT. It is worth noting that, without class-discriminative region identification, the focus stage is executed on the entire image, introducing a large number of redundant tokens and significantly increasing computational cost. In contrast, our class-discriminative region identification ensures that the focus stage focuses only on the class-discriminative region, drastically reducing the number of tokens and thus improving efficiency without compromising accuracy.

\textbf{Influence of $\beta$ and $\alpha$.} Table~\ref{tab:tab4} and Table~\ref{tab:tab5} show the effect of $\beta$ and $\alpha$ on the accuracy of the LF-ViT focus stage, respectively. Here, $\beta$ represents the weight from shallow encoders when calculating the GCA. $\alpha$ denotes the number of tokens selected from the tokens representation of the class-discriminative region. In this paper, we choose the setting that achieves the best performance $\beta$ = 0.99 and $\alpha$ = 0.88 as default values.

\textbf{Class-discriminative Regions Identification and Localization.} Four variants are developed to replace our class-discriminative region identification: (1) Negative neighborhood global class attention (negative NGCA): which selects regions with the smallest neighboring area based on GCA to serve as class-discriminative regions. (2) Maximum global class attention (maximum GCA): which selects the region of size $m$ containing the tokens with the highest GCA as the class-discriminative region. (3) Minimum global class attention (minimum GCA): which selects the region of size $m$ containing the tokens with the smallest GCA as the class-discriminative region. (4) Random: which selects randomly regions of size $m$ to serve as class-discriminative regions. 

In Table~\ref{tab:tab7}, we remove the early-exiting design, the class-discriminative region size $m$ = 5, and show the performance of four class-discriminative region identification methods in both the localization inference stage and focus inference stage. We find that the negative NGCA region identification method yielded the worst results, as it selected non-class-discriminative regions, leading to a significant drop in accuracy during the focus inference stage due to the loss of class-discriminative characteristics. 
On the other hand, the maximum GCA region method has the highest accuracy among the compared alternatives because the maximum GCA method includes the regions around it, which are usually important for correctly recognizing the image as well. In comparison to these methods, our NGCA mechanism always selects regions with the maximum NGCA, resulting in the highest accuracy overall. Therefore, we chose NGCA as our default class-discriminative region identification method.

\begin{table}[t]

\centering
\begin{tabular}{ccc}
\hline
     \multirow{2}{*}{  Ablation}  & \multicolumn{2}{c}{ Top-1 Acc.(\%)}\\
     
     & location & focus \\
\hline
negative NGCA & 75.3 & 79.7 \\
maximum GCA & 75.8 & 80.2\\
minimum  GCA & 75.4 & 79.8\\
     random & 75.7 & 80.0\\
    \hline
    \textbf{Ours} & \textbf{76.1} & \textbf{80.8} \\
    \hline

\end{tabular}
\caption{Performance comparison between our class-discriminative region recognition and its variants. NGCA and GCA mean neighborhood global class attention and global class attention respectively.}
\label{tab:tab7}
\end{table}

\begin{table}[t]

\centering
\begin{tabular}{ccc}
\hline
     \multirow{2}{*}{  Ablation}  & \multicolumn{2}{c}{ Top-1 Acc.(\%)}\\
     
     & location & focus \\
\hline
CE + CE & 76.1 & 80.4 \\

    \hline
    \textbf{CE + KL(Ours)} & \textbf{76.1} & \textbf{80.8} \\
    \hline

\end{tabular}
\caption{Performance comparison between different loss functions.}
\label{tab:tab6}
\end{table}
\begin{figure*}[t]
\centering
\includegraphics[width=1.80\columnwidth]{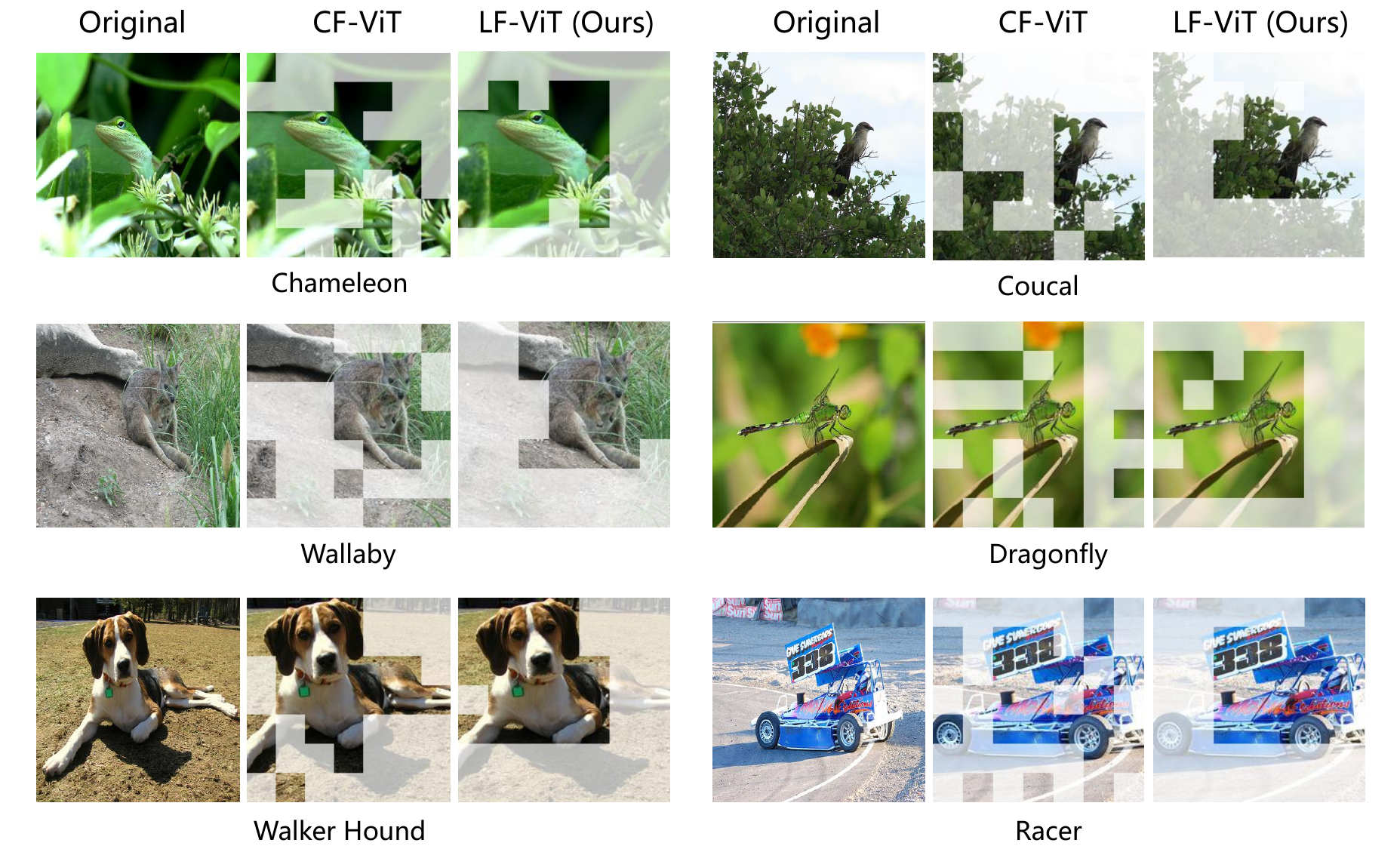}
\caption{Comparison of our LF-ViT and CF-ViT \cite{cf_vit} visualizations. We visualize the regions selected by our LF-ViT class-discriminative region identification or CF-ViT informative region identification (grey boxes) to indicate the uninformative patches. We find that our method can obviously better focus the objects of interest. The visualized results here are randomly selected images that are correctly recognized by both the LF-ViT and CF-ViT methods.}
\label{fig:fig7}
\end{figure*}

\begin{figure}[t]
\centering
\includegraphics[width=1.0\columnwidth]{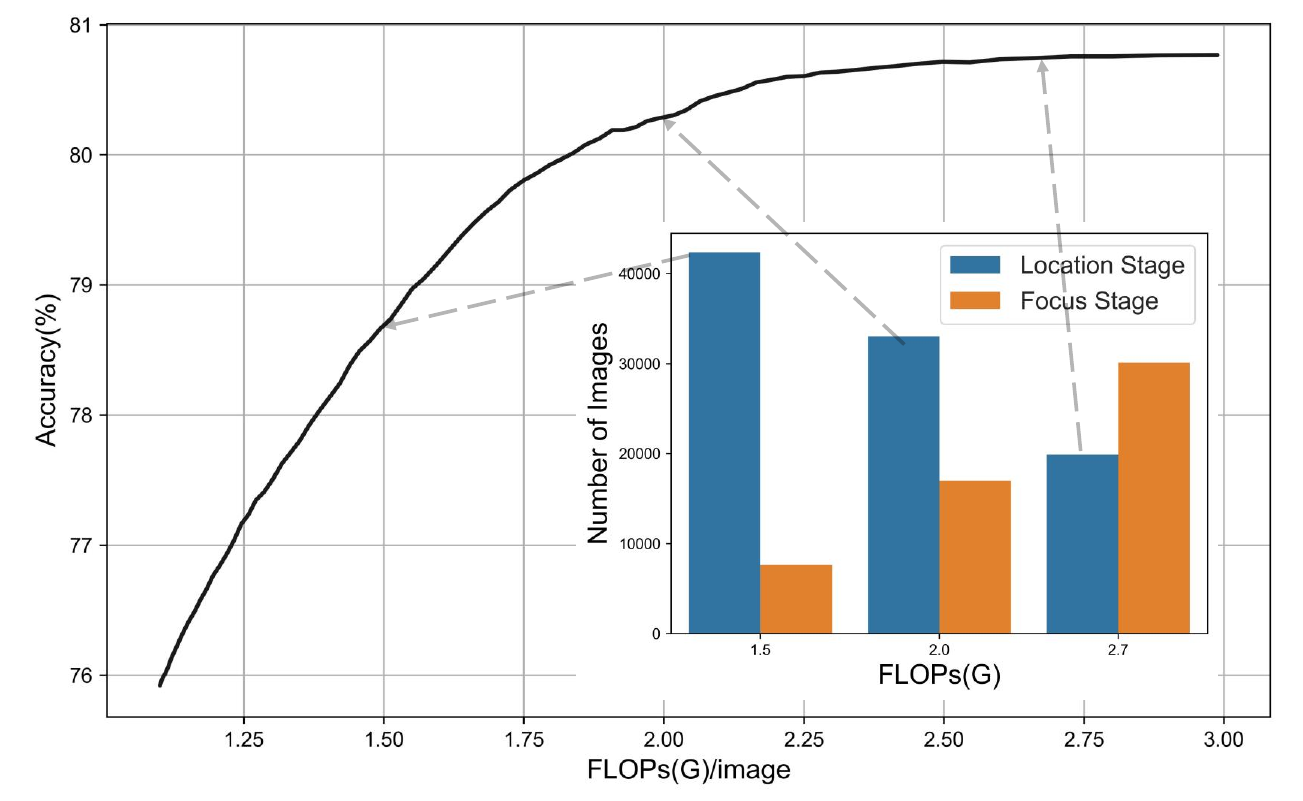}
\caption{The number of images correctly classified by LF-ViT in the localization and focus stages.}
\label{appedixfig:fig2}
\end{figure}

\textbf{Influence of Loss Function.} During the training of LF-ViT using Eq.~12, for the output of the focus stage, we use the Cross-Entropy (CE) loss function and supervise training it with ground truth (GT) labels. For the output of the localization stage, we use the Kullback-Leibler (KL) loss function and supervise training it with the output of the focus stage. To further investigate the impact of different loss functions on LF-ViT's performance, we also employ the CE loss function from Eq.~1 and use GT labels to supervise the localization inference outputs of LF-ViT:

\begin{equation}
    \hat{\mathcal{L}_{cls}} = CE(\mathbf{p}_f; \mathbf{y}) + KL(\mathbf{p}_l; \mathbf{y})
\end{equation}

Table~\ref{tab:tab6} shows the impact of different loss functions on the performance of LF-ViT. The results using CE + CE as the loss function for training in the localization stage have the same accuracy as those using CE + KL, but the accuracy drops by 0.4\% in the focus inference stage. Therefore, we choose CE + KL as the default training loss function.

\subsection{Visualization and Statistical Analysis}
As shown in Fig.~\ref{fig:fig7}, we visualize examples of LF-ViT correctly recognizing images in the focus stage. Also, we visualize examples of CF-ViT \cite{cf_vit} correctly recognizing the same images in the fine-grained inference stage for comparison. For a better illustration, we only visualize informative regions if images are recognized in the focus stage (fine inference stage of Cf-ViT).

By observing the results in the figures, it is evident that LF-ViT consistently focuses its attention on the regions of interest during the focus inference stage, while CF-ViT's fine-grained inference stage covers the entire image and involves more tokens, resulting in increased computational cost. This indicates that LF-ViT efficiently reduces unnecessary computations by concentrating attention on key regions, thus improving computational efficiency while maintaining accuracy. In contrast, CF-ViT's fine-grained inference strategy introduces significant redundant computations, leading to performance degradation and potential limitations on resource-constrained devices. Therefore, our LF-ViT achieves a better balance between model performance and efficiency, demonstrating outstanding performance and practicality.

In Fig.~\ref{appedixfig:fig2}, we explore the impact of varying $\eta$ on the accuracy at different FLOPs and simultaneously analyzed the number of recognize images during the localization and focus inference stages of LF-ViT at different accuracy levels. A smaller value of $\eta$ leads to a decrease in LF-ViT's accuracy, primarily because more images are terminated during the localization inference stage. Conversely, a larger value of $\eta$ results in more images being sent to the focus inference stage, leading to increase accuracy. These findings demonstrate that LF-ViT offers a certain level of flexibility in balancing accuracy and computational efficiency. By adjusting the value of $\eta$, the model's performance and speed can optimization according to specific application requirements.

\section{Conclusion}
This paper addresses the issue of redundant input tokens in the acceleration of ViT models for processing high-resolution images, primarily considering the images' spatial redundancy. We introduce the Localization and Focus Vision Transformer (LF-ViT). Its inference operates in two main stages: localization and focus. Initially, the input image is down-sampled during the localization stage to distinguish ``easy" images and pinpoint the class-discriminative regions in ``hard" images. If this stage cannot make a high-confidence prediction, the focus stage steps in, concentrating on the localized class-discriminative areas. Our comprehensive experiments reveal that LF-ViT strikes an improved balance between performance and efficiency.

\section{Acknowledgments}
This work was supposed by NSFC under Grant No. 62072137.

\bibliography{aaai24}

\end{document}